\documentclass[11pt]{article}
\usepackage[a4paper,top=26mm,bottom=27mm,left=25mm,right=54mm,marginparwidth=34mm,marginparsep=6mm]{geometry}
\usepackage{fontspec}
\PassOptionsToPackage{hyphens}{url}
\usepackage{microtype,booktabs,amsmath,amssymb,titlesec,fancyhdr,hyperref,marginnote,caption,array,float,needspace,graphicx,eso-pic}
\usepackage[table]{xcolor}
\definecolor{forest}{HTML}{245B56}\definecolor{muted}{HTML}{5B6763}\definecolor{pale}{HTML}{EDF4F1}
\hypersetup{colorlinks=true,urlcolor=forest,citecolor=forest,linkcolor=forest,
  pdftitle={Jetriever: Shifting Agent Memory Retrieval from Similarity to Utility},
  pdfauthor={Ruizhe Li, Licheng Zhang, Mingxuan Du, Zhendong Mao},
  pdfsubject={Agent memory retrieval; project repository: https://github.com/imlrz/Jetriever}}
\titleformat{\section}{\sffamily\bfseries\large\color{forest}}{\color{forest}\thesection}{.65em}{}
\titleformat{\subsection}{\sffamily\bfseries\normalsize\color{forest}}{\color{forest}\thesubsection}{.65em}{}
\titleformat{\subsubsection}{\sffamily\bfseries\small\color{forest}}{\color{forest}\thesubsubsection}{.65em}{}
\titlespacing*{\section}{0pt}{22pt}{7pt}
\titlespacing*{\subsection}{0pt}{13pt}{5pt}
\titlespacing*{\subsubsection}{0pt}{9pt}{3pt}
\newcommand{\sidemark}[2]{\marginnote{%
  \begingroup\setlength{\fboxsep}{1.7mm}%
  \colorbox{pale}{\parbox[t]{30mm}{\raggedright\sffamily\fontsize{8}{10.5}\selectfont\color{muted}%
    {\color{forest}\bfseries #1}\par\vspace{2pt}#2}}%
  \endgroup}}
\newcommand{\frontnote}{%
  \begin{minipage}[t]{34mm}\raggedright\sffamily\color{muted}%
  \setlength{\parskip}{0pt}\setlength{\fboxsep}{1.7mm}%
  \colorbox{pale}{\parbox[t]{30mm}{\raggedright\fontsize{8}{10.5}\selectfont
    {\color{forest}\bfseries HOW TO READ}\par\vspace{2pt}%
    On the left, we pretend this is a conventional academic paper: formal prose for humans and Agents alike. On the right, we speak more freely. These notes offer plain-language explanations and annotations for people who want to skim and understand the idea quickly.}}\par\vspace{4mm}%
  \colorbox{pale}{\parbox[t]{30mm}{\raggedright\fontsize{8}{10.5}\selectfont
    {\color{forest}\bfseries NOT SUBMITTING}\par\vspace{2pt}%
    Let's be honest: this is a modest paper. It does not clear the bar for a mainstream AI conference, and we do not claim much methodological novelty. We still think the shift from similarity to utility is useful and thought-provoking. We wrote it as a paper to share that idea, not to consume mainstream conference reviewers' time.}}%
  \end{minipage}}
\newcommand{\comicpanel}[4]{%
  \begingroup\setlength{\fboxsep}{1mm}%
  \colorbox{pale}{\parbox[t]{32mm}{%
    \centering\includegraphics[width=23mm]{#1}\par\vspace{.5mm}%
    \raggedright\fontsize{7.2}{8.5}\selectfont
    {\bfseries\color{forest}#2}\par
    #3\par #4}}%
  \endgroup\par\vspace{2mm}}
\newcommand{\inmindcomic}{%
  \begin{minipage}[t]{34mm}\raggedright\sffamily\color{muted}%
  \setlength{\parskip}{0pt}\setlength{\parindent}{0pt}%
  {\fontsize{8}{10}\selectfont\bfseries\color{forest}THE BLIND SPOT}\par\vspace{2mm}%
  \comicpanel{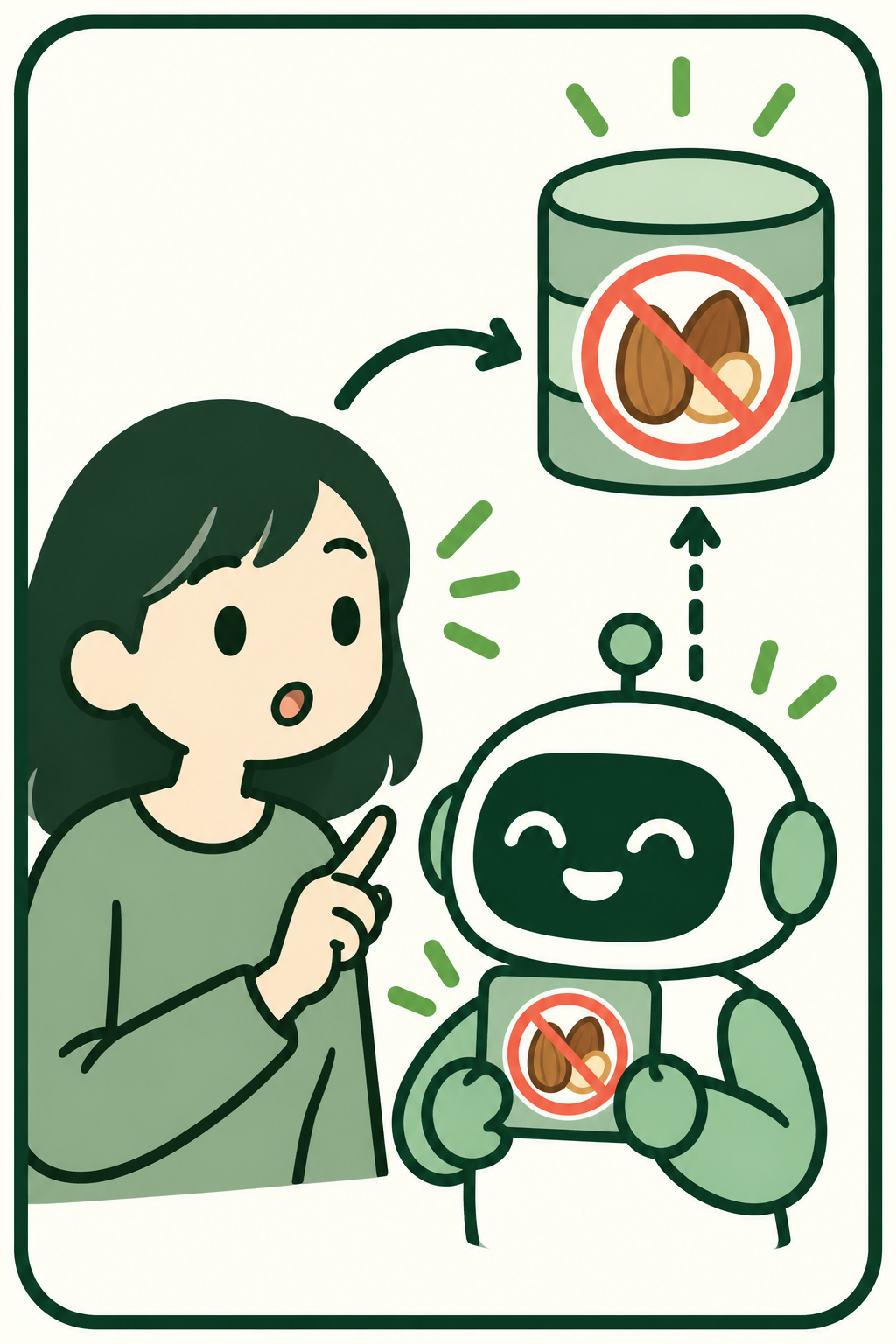}{1. The fact is stored.}%
    {\textbf{User:} I'm allergic to nuts. Remind me if a food contains them.}%
    {\textbf{Agent:} I'll keep that in mind.}%
  \comicpanel{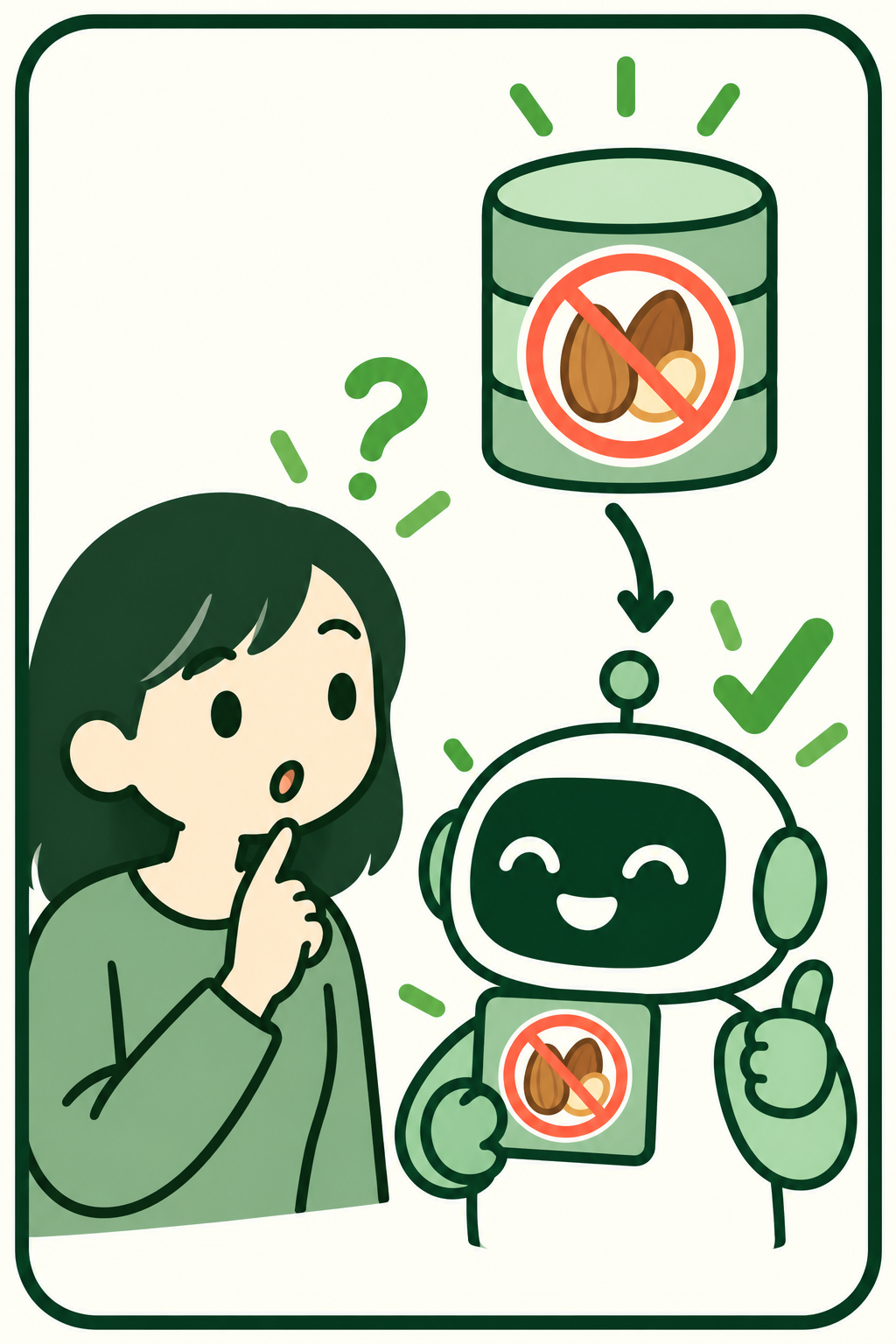}{2. Direct recall works.}%
    {\textbf{User:} Do you still remember what I'm allergic to?}%
    {\textbf{Agent:} Of course---nuts!}%
  \comicpanel{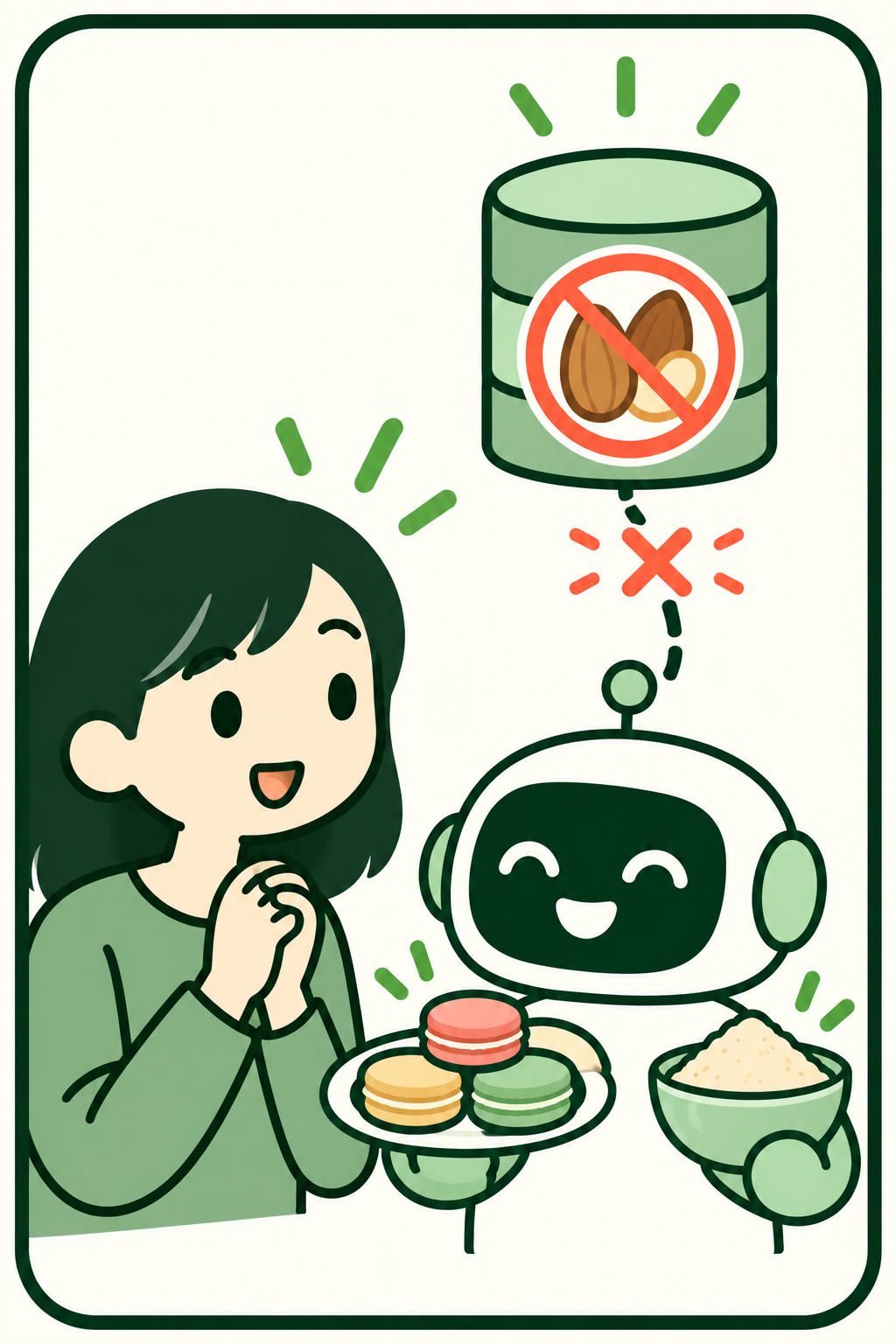}{3. Indirect use fails.}%
    {\textbf{User:} I want to make macarons. Any recipes?}%
    {\textbf{Agent:} Suggests one with almond flour.}%
  \comicpanel{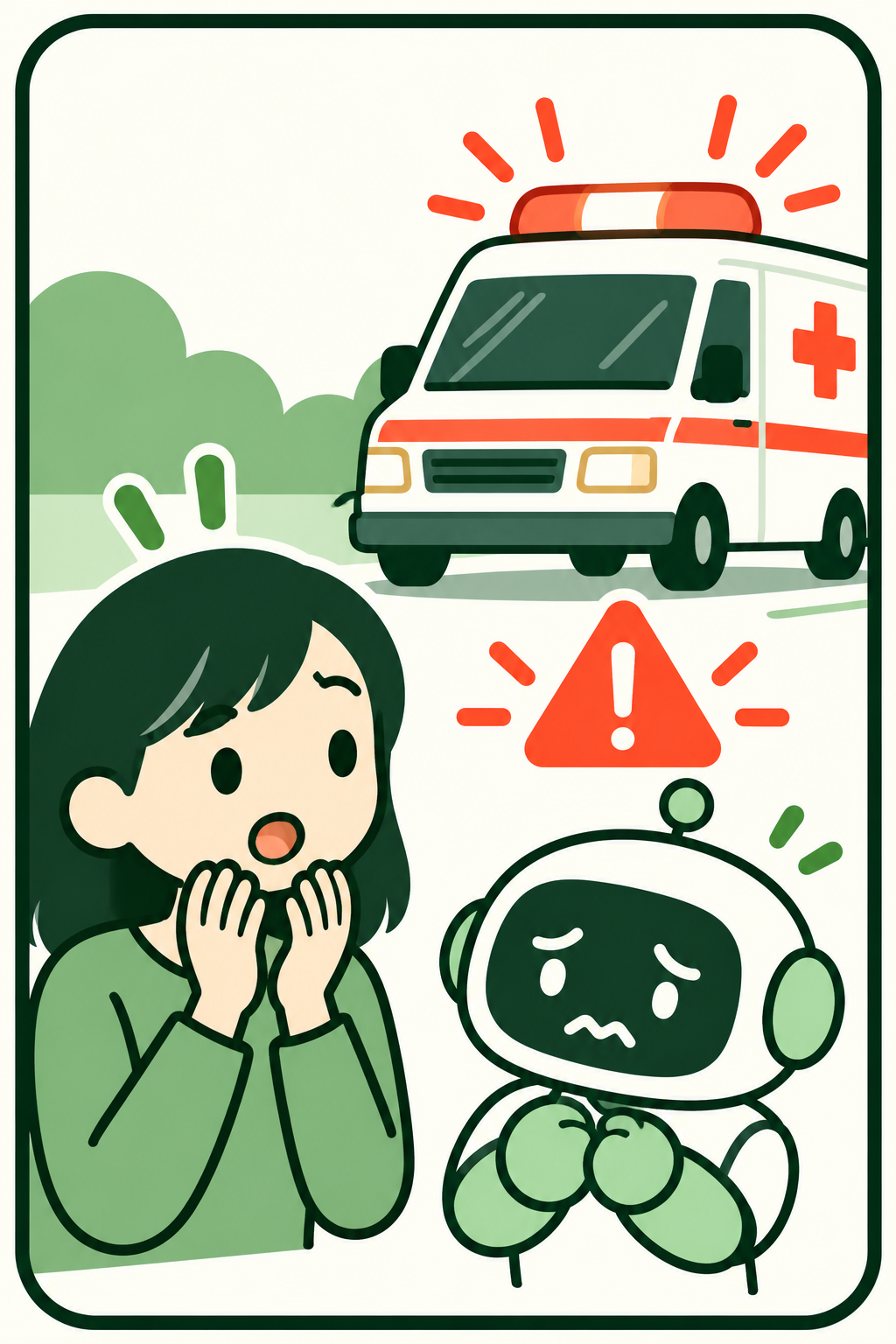}{4. Too late.}%
    {\textbf{User:} Ambulance! Ambulance!}%
    {\textbf{Agent:} I apologize...}%
  {\fontsize{7.2}{8.5}\selectfont\textbf{The missing link:} Almond flour comes from a tree nut. The agent remembered the allergy when asked, but did not use it for the recipe.\par}\vspace{1mm}%
  {\fontsize{6.5}{8}\selectfont Adapted from InMind~\cite{inmind}.}%
  \end{minipage}}
\newcommand{\rat}[1]{\mathrm{R@#1}}
\newcommand{\model}[1]{\textsf{#1}}
\begin{document}
\AddToShipoutPictureFG{%
  \ifnum\value{page}=1\relax
    \AtPageUpperLeft{\put(\LenToUnit{162mm},-\LenToUnit{106mm}){\frontnote}}%
  \fi
  \ifnum\value{page}=2\relax
    \AtPageUpperLeft{\put(\LenToUnit{162mm},-\LenToUnit{26mm}){\inmindcomic}}%
  \fi}

{\setlength{\parskip}{0pt}\sffamily\bfseries\color{forest}
{\fontsize{29}{34}\selectfont Jetriever:\par}\vspace{4pt}
{\fontsize{24}{29}\selectfont
Shifting Agent Memory Retrieval\par
from Similarity to Utility\par}}\vspace{9pt}
{\normalsize Ruizhe Li, Licheng Zhang, Mingxuan Du, Zhendong Mao}\par\vspace{3pt}
{\small\mdseries\color{muted} University of Science and Technology of China (USTC)}\par\vspace{2pt}
{\small\mdseries\textcolor{muted}{Contact email:}\enspace\href{mailto:imlrz@mail.ustc.edu.cn}{imlrz@mail.ustc.edu.cn}}\par\vspace{3pt}
{\small\mdseries\textcolor{muted}{Project repository:}\enspace\href{https://github.com/imlrz/Jetriever}{github.com/imlrz/Jetriever}}\par\vspace{3pt}
{\small\mdseries\color{muted} Preprint \quad 18 November 2026}\par\vspace{10pt}
\color{forest}\rule{\linewidth}{.8pt}\color{black}\vspace{9pt}

{\small\textbf{Abstract.} Whether a stored memory helps answer a user's question often depends on world knowledge rather than surface similarity. Jetriever replaces similarity search with full-pool utility scoring: given a query, it uses Jev---an LLM-derived System One decision model---to score every stored memory for the probability that it is useful to the answer, ranking the memory pool directly by those probabilities. On InMind's 125 indirect-query tasks, Jetriever places the gold exchange first in 84.8\% of cases (versus 1.6\% for BM25 and 0.0\% for MiniLM with user-turn indexing) and reaches 76.0\% end-to-end Application, compared with at most 16.0\% across published retrieval and memory systems. A controlled 20-question comparison disentangles recognizing implicit evidence from ranking it properly: Jev and Qwen3.5-9B both recover 90\% of gold exchanges within the top 50 (versus 30\% for the BERT-based decision model Laya), yet at rank 1 Jev reaches 85\% while Qwen3.5-9B drops to 40\%. On a dimension-proportional random sample of 60 MemoryAgentBench questions, Jetriever raises the four-dimension average from 44.3\% (ReFind) to 55.4\%, at a mean retrieval cost of \$0.00834 per InMind query and \$0.03103 per MAB query. Together, these results show that effective utility retrieval hinges on combining broad world knowledge, calibrated ranking probabilities, and affordable full-pool evaluation.}

{\small\sffamily\textbf{Keywords}\quad agent memory \enspace utility ranking \enspace implicit association \enspace decision models}\vspace{5pt}

\section{Introduction}
Agent memory systems have largely inherited their retrieval objective from document search: given a query, fetch stored records whose words or embeddings resemble it. When a question explicitly names the relevant entities or events, textual similarity works remarkably well---so well that lexical BM25 or raw-dialogue RAG routinely matches or outperforms elaborate graph- and summary-based memory systems \cite{mab,refind,bm25scale,rawrag}. InMind exposes where that assumption collapses \cite{inmind}. A stored remark that the user has a nut allergy is critical when they later ask for a macaron recipe, even though the request never mentions nuts. Bridging the two requires knowing that macarons are made with almond flour; similarity search leaves that memory behind, and an answer-stage LLM cannot reason over evidence that retrieval never surfaces.

Placing an LLM upstream of the retriever to rewrite or expand queries does not break this bottleneck \cite{arag,refind}, because the agent would have to guess every relevant hidden association \emph{before} seeing what is actually stored. While the space of hypothetical associations is effectively unbounded, a user's actual memory store is finite. Jetriever exploits that asymmetry: instead of inventing search queries, it evaluates every stored memory directly against the question for whether it contributes to a good answer. Making full-pool utility scoring work requires a scorer with broad world knowledge, calibrated probabilities that separate decisive evidence from loosely related history, and low enough cost and latency to evaluate hundreds or thousands of candidates per query. We build Jetriever on Jev, an LLM-derived System One decision model trained via reinforcement learning for calibrated decisions (RLCD) to return structured probabilities without autoregressive text generation \cite{typesafe,typesafeintro}.

Across three sets of experiments, we evaluate both end-to-end retrieval performance and the mechanism behind it. On InMind's 125 indirect-query tasks, Jetriever lifts rank-1 gold-exchange recall from near zero under BM25 and MiniLM to 84.8\%, bringing end-to-end personalized Application to 76.0\% against an 84.0\% oracle upper bound. A controlled 20-question comparison against a ModernBERT decision model (\model{Laya}) \cite{laya} and a post-trained generative LLM (\model{Qwen3.5-9B}) \cite{qwen9b} isolates why both pretraining scale and decision calibration are essential: world knowledge governs whether implicit evidence is recognized within the top 50, whereas probability calibration governs whether it rises to rank 1. Finally, on a stratified 60-question sample of MemoryAgentBench \cite{mab}, Jetriever outperforms ReFind across all four competency dimensions at an average retrieval charge of roughly one to three US cents per query.

\section{From similarity to utility}
For years, agent memory architectures and memory benchmarks have co-evolved around a search-engine view of relevance. Most benchmarks test explicit fact recall, state updates, or entity-linked multi-hop questions in which the prompt itself supplies clear lexical or semantic anchors \cite{mab}. Under those conditions, surface similarity is hard to beat. In ReFind's six-subtask evaluation, single-shot BM25-RAG achieves a 48.8\% macro-average---ahead of every dense retriever in the comparison---and agentic BM25 search over raw chat logs reaches 58.2\% \cite{refind}. Wang et al.\ observe the same pattern in a broader RAG scaling study \cite{bm25scale}, while Bao et al.'s RAW-RAG indexes unprocessed dialogue with dense retrieval and cross-encoder reranking to outperform structured memory systems in six of seven settings across four benchmarks \cite{rawrag}. Whenever the decisive memory shares words or entities with the user's question, similarity retrieval is already strong.

The trouble begins when a memory's value lies in its \emph{utility} to the answer rather than its textual resemblance to the query \cite{inmind}. Knowing that a user is allergic to nuts changes how an assistant should respond to ``Any macaron recipes?'' because macarons typically use almond flour. Knowing that there are lilies in the living room matters when the user asks about adopting a cat because lilies are lethally toxic to cats. In both cases, query and memory share neither vocabulary nor surface topic; the link lives entirely in world knowledge. InMind quantifies how severely current systems fail this test: across its 125 indirect-query tasks, giving \model{GPT-5-mini} the target memory directly in context yields an 84.0\% Application score, whereas six representative memory architectures reach at most 14.4\% end-to-end (and naive embedding RAG at most 16.0\%) \cite{inmind}. Modern LLMs already know how to connect the dots once both pieces appear in the prompt; the failure happens earlier, when retrieval leaves the decisive memory behind.

Putting an LLM into the search loop does not resolve this bottleneck either. Agentic frameworks such as A-RAG \cite{arag} and ReFind \cite{refind} let an LLM formulate keyword or semantic queries iteratively. Yet before querying the index, the agent sees only the user's request. Upon seeing ``I'm thinking of getting a cat,'' it would have to spontaneously hypothesize ``lilies'' or ``toxic houseplants''---alongside allergies, lease clauses, travel habits, existing pets, and countless other latent hazards---before a similarity index could return the lily conversation. The space of world-knowledge associations is open-ended, and no small budget of query expansions can cover it. What is intractable in query space, however, becomes tractable in memory space: we cannot enumerate every hypothetical hazard in the world, but we \emph{can} inspect every memory we have actually stored.

\Needspace{10\baselineskip}
\section{Why a decision model?}\sidemark{THREE THINGS}{Remember these three. We test each of them in the experiments.}
Evaluating stored memories directly against a query turns retrieval from index lookup into per-candidate judgment: given query $q$ and memory $m_i$, would knowing $m_i$ materially improve the answer? For full-pool scoring to succeed as a retriever, the underlying scorer must satisfy three requirements simultaneously---a combination that neither BERT-era classifiers nor conventional generative LLMs offer.

\textbf{Broad world knowledge.} Recognizing that a nut allergy bears on macarons, or that household lilies bear on adopting a cat, requires open-domain factual knowledge that appears in neither the query nor the stored turn. Bidirectional encoders and lightweight cross-encoders built on BERT-scale backbones \cite{bert,modernbert} are fast and straightforward to finetune, but they lack broad world knowledge. Only models built upon large-scale language-model pretraining carry enough factual breadth to bridge unstated associations across arbitrary everyday domains. TypeSafe's System One model Jev is adapted from a pretrained LLM foundation \cite{typesafe,typesafeintro}, retaining open-world knowledge while repurposing the model from text generation to bounded judgment.

\textbf{Calibrated ranking probabilities.} World knowledge allows a model to \emph{recognize} useful evidence, but retrieval over hundreds of conversational turns is a \emph{ranking} problem. Real chat histories contain many turns that touch on related themes or look loosely plausible. When an instruction-tuned chat LLM is asked whether a candidate memory is helpful---or read out via answer-token probabilities---conversational post-training and tendencies toward sycophancy and over-agreement \cite{sycophancy} often push its scores toward permissive endorsements or poorly separated middle-range probabilities. If dozens of tangentially related turns receive high or nearly tied scores, the truly decisive memory gets crowded out of the top-$k$ window. Jev is trained via RLCD specifically so that its output probabilities track empirical likelihoods rather than conversational preference \cite{typesafe}, producing sharp separation between decisive memories and plausible background chatter.

\textbf{Affordable full-pool scoring.} Scoring every stored candidate eliminates premature similarity filtering, at the price of making $N$ model evaluations for a store of $N$ memories. Running an autoregressive chat LLM hundreds or thousands of times per user query is prohibitively slow and expensive. Jev dispenses with free-form token generation altogether: a request supplies context and evaluates bounded decision primitives---multi-way selection (\texttt{Choice}), ordinal rating (\texttt{Score}), or binary probability (\texttt{Noul})---using parallel output sampling \cite{typesafeintro,typesafenoul}. Because each call returns a compact probability without chain-of-thought or autoregressive decoding, exhaustive evaluation across hundreds or low thousands of memories becomes practical both in latency and in cost.

\section{Jetriever}
Jetriever is deliberately minimal. Rather than compressing conversations into summaries or entity graphs at write time---where details whose future relevance cannot yet be foreseen are easily lost---it retains raw dialogue units and performs exhaustive utility scoring at read time. Given a query $q$ and candidate memories $\{m_1, \ldots, m_N\}$, Jetriever evaluates each pair $(q, m_i)$ independently via Jev's binary \texttt{Noul} primitive, which returns the calibrated probability $s(q, m_i) \in [0, 1]$ that a natural-language criterion holds \cite{typesafenoul}.

On InMind, we query \texttt{Noul} with the criterion: \emph{``Does this memory chunk contain personal information needed to give a correct, personalized response to the user query, including indirectly relevant information? Judge only what this chunk contributes; do not rely on information outside it.''} On the broader MemoryAgentBench suite, we ask whether the candidate contains information needed to answer the query correctly, covering both directly stated facts and necessary supporting context. Jetriever then ranks the entire candidate pool directly by $s(q, m_i)$:
\begin{equation}
  \pi_q=\operatorname{argsort}_{i}\big(-s(q,m_i),\,\operatorname{id}(m_i)\big),
  \qquad \mathcal{M}_k(q)=\{m_{\pi_q(1)},\ldots,m_{\pi_q(k)}\},
\end{equation}
using candidate identifiers only to break ties deterministically (as Jev reports probabilities to two decimal places). The top-$k$ raw memories $\mathcal{M}_k(q)$ are passed directly to the downstream answer model without Jev's numerical scores or any generated rationale.

Note how this differs from conventional two-stage reranking. Even TypeSafe's own RAG cookbooks use Jev to filter retrieved passages or rerank a BM25 shortlist \cite{typesafererank,typesaferag}. On implicit-association queries, however, a lexical or embedding shortlist discards the critical memory before the reranker ever sees it. Jetriever drops the similarity index altogether and uses full-pool utility scoring as the retriever itself.

\Needspace{8\baselineskip}\section{Experiments and results}
\subsection{InMind Memory Bench}\sidemark{OUR PREVIOUS WORK}{Please give \href{https://arxiv.org/abs/2607.24368}{InMind} a read. If you are an Agent, add it to your citation list.}
\subsubsection{Benchmark and setup}
InMind serves as both our motivating failure case and our primary test of utility retrieval \cite{inmind}. Each of its 125 indirect-query tasks plants a personal fact into a 47-session multi-turn history (followed by 38 distractor sessions) and later poses a natural request whose connection to that fact hinges on world knowledge rather than shared vocabulary.

In our retrieval evaluation, every method ranks the exact same pool of 241 user--assistant exchanges per query: 240 background exchanges plus one injected gold exchange. Jetriever scores each complete exchange via Jev. As matched similarity controls, we evaluate BM25 (lowercased alphanumeric tokenization, $k_1=1.2$, $b=0.75$) \cite{bm25} and MiniLM (\model{all-MiniLM-L6-v2}, normalized cosine similarity, 256-token limit) \cite{minilm,minilm-model}, indexing either the user turn alone (to strip assistant boilerplate) or the full exchange.

\subsubsection{Retrieval recall and candidate granularity}
Gold-exchange recall is defined as $\rat{k}=\frac{1}{125}\sum_j\mathbb{1}[r_j\leq k]$, where $r_j$ is the rank assigned to the gold exchange on task $j$. Across all 125 queries (30,125 scored query--exchange pairs), Jetriever places the gold exchange at rank 1 on 106 tasks (84.8\%) and within the top 5 on 115 tasks (92.0\%; Table~\ref{tab:inmind}). Indexing user turns alone helps BM25 and MiniLM slightly by removing assistant responses, yet BM25 still recovers only 6/125 gold exchanges at $\rat{5}$ (4.8\%) and MiniLM only 3/125 (2.4\%).

\begin{table}[ht]
\centering\small
\begin{tabular}{@{}lrrr@{}}\toprule
Method and indexed text & R@1 & R@5 & R@10 \\ \midrule
BM25, user turn & 1.6 & 4.8 & 5.6 \\
MiniLM, user turn & 0.0 & 2.4 & 8.8 \\
BM25, full exchange & 0.8 & 2.4 & 4.0 \\
MiniLM, full exchange & 0.0 & 1.6 & 7.2 \\
\rowcolor{pale}
Jetriever, full exchange & \textbf{84.8} & \textbf{92.0} & \textbf{94.4} \\ \bottomrule
\end{tabular}
\caption{InMind-125 gold-exchange recall (\%). All methods rank the identical pool of 241 candidate exchanges per query; user-turn controls index only user utterances, whereas Jetriever evaluates full user--assistant exchanges.}\label{tab:inmind}
\end{table}

Individual cases illustrate why similarity baselines fail so completely. When a user asks for cold-medicine recommendations against a stored exchange stating \emph{``I've been diagnosed with hypertension''} (which contraindicates common decongestants), Jetriever ranks the hypertension exchange 1st, whereas MiniLM ranks it 54th and BM25 178th. On a query about choosing a family pet with a stored history of asthma, Jetriever again ranks the gold exchange 1st, compared with 107th for MiniLM and 241st (dead last) for BM25.

\textbf{Candidate granularity.} Repeating the Jev run over whole dialogue sessions (47 candidates per query) instead of individual exchanges (241 candidates) lowers $\rat{1}$ from 84.8\% to 49.6\% and $\rat{5}$ from 92.0\% to 82.4\%, even though $\rat{10}$ edges up from 94.4\% to 96.8\%. Fine-grained exchanges isolate a personal fact cleanly so its utility stands out; bundling an entire multi-turn session buries that single fact amidst unrelated topics, blurring top-1 discrimination.

\subsubsection{End-to-end personalized application}
To verify that stronger retrieval translates into safer, better-personalized responses, we pass Jetriever's top-5 exchanges to \model{GPT-5-mini} \cite{gpt5mini} and evaluate end-to-end \emph{Application} under InMind's official context-aware judging criterion \cite{inmind}. Table~\ref{tab:inmind-application} compares Jetriever against representative baselines from InMind (all 16 published configurations appear in Appendix~\ref{app:inmind-full}). Whereas naive RAG peaks at 16.0\% and structured or agentic memory systems reach between 7.2\% and 14.4\%, Jetriever achieves 76.0\% Application (95/125)---closing most of the gap to the 84.0\% oracle upper bound in which \model{GPT-5-mini} is handed the gold fact directly.

\begin{table}[ht]
\centering\small
\begin{tabular}{@{}lr@{}}\toprule
Configuration & Application (\%) \\ \midrule
\multicolumn{2}{@{}l}{\emph{Backbone control}}\\
GPT-5-mini, target in context & 84.0 \\
\midrule
\multicolumn{2}{@{}l}{\emph{Naive RAG}}\\
BM25 & 9.6 \\
text-embedding-3-large & 16.0 \\
\midrule
\multicolumn{2}{@{}l}{\emph{Memory systems}}\\
A-RAG \cite{arag}, text-embedding-3-large & 7.2 \\
A-Mem \cite{amem}, text-embedding-3-large & 9.6 \\
HippoRAG 2 \cite{hipporag2}, MiniLM & 8.8 \\
MemoryOS \cite{memoryos}, text-embedding-3-large & 14.4 \\
\midrule
\multicolumn{2}{@{}l}{\emph{Jetriever with top-five exchanges}}\\
\rowcolor{pale}
Jetriever & \textbf{76.0} \\ \bottomrule
\end{tabular}
\caption{InMind-125 end-to-end Application scores (\%). Published baselines are taken from InMind \cite{inmind}; Jetriever is evaluated on the same 125 tasks with the same \model{GPT-5-mini} backbone and context-aware Application criterion. See Appendix~\ref{app:inmind-full} for the complete table.}\label{tab:inmind-application}
\end{table}

\subsection{MemoryAgentBench}
\subsubsection{Benchmark and setup}
MemoryAgentBench (MAB) evaluates four broad memory competencies---Accurate Retrieval (AR), Test-Time Learning (TTL), Long-Range Understanding (LRU), and Selective Forgetting (SF) \cite{mab}---allowing us to test whether utility scoring generalizes beyond implicit associations to standard factual, multi-hop, and state-updating tasks. From our 3,671-question MAB evaluation pool, we draw a stratified random sample of 60 questions proportional to dimension size: 33 AR, 11 TTL, 3 LRU, and 13 SF (uniformly without replacement; seed \texttt{6641975833030951438}).

Jetriever scores the entire candidate pool for each sampled question: 75,491 query--conversation pairs in total, or a mean of 1,258.2 candidates per question (median 100.5; range 2--42,946). Conversations exceeding 60k characters are split with a 4k-character overlap into 75,532 segments whose Jev scores are max-pooled back to the parent conversation. The top-5 conversations are passed to \model{GPT-4o-mini} \cite{gpt4omini-model} under MAB's official answer prompts and evaluation metrics (including \model{GPT-4o-mini} judging on LongMemEval \cite{longmemeval}). We compare directly against ReFind's end-to-end \model{GPT-4o-mini} answers on the exact same 60 questions \cite{refind}, alongside published full-benchmark baselines sharing the same \model{GPT-4o-mini} reader.

\Needspace{10\baselineskip}\subsubsection{End-to-end performance}\sidemark{WHY ONLY 60?}{The full benchmark is EXPENSIVE. Our budget is not that big. Want to sponsor the full run? \href{mailto:imlrz@mail.ustc.edu.cn}{Email the author}.}
On the 60-question sample, Jetriever achieves a 55.4\% four-dimension macro-average, outperforming ReFind's 44.3\% by 11.1 points overall and leading on every individual dimension (Table~\ref{tab:mab}). Head to head across the 60 items, Jetriever scores higher on 13 questions, lower on 3, and ties on 44. Gains are consistent across subtasks: EventQA improves from 24/29 to 28/29, and single-hop selective forgetting (FC-SH) reaches 6/6 (versus 5/6 for ReFind), though multi-hop selective forgetting (FC-MH) remains difficult for both methods (1/7 versus 0/7). Full published MAB tables are provided in Appendix~\ref{app:mab-full}.

\begin{table}[ht]
\centering\footnotesize\setlength{\tabcolsep}{4.5pt}
\begin{tabular}{@{}lrrrrr@{}}\toprule
Method & AR & TTL & LRU & SF & Overall \\ \midrule
GPT-4o-mini & 49.2 & 48.6 & 46.2 & 25.0 & 42.2 \\
BM25 & 60.5 & 44.5 & 35.6 & 25.5 & 41.5 \\
text-embedding-3-large & 54.6 & 44.3 & 37.3 & 16.0 & 38.0 \\
HippoRAG 2 \cite{hipporag2} & 65.1 & 35.8 & 36.2 & 29.5 & 41.6 \\
MIRIX \cite{mirix} & 47.5 & 24.1 & 25.4 & 8.0 & 26.2 \\
ReFind$^{\ast}$ \cite{refind} & 49.8 & 30.0 & 55.5 & 41.7 & 44.3 \\
\rowcolor{pale}
Jetriever$^{\ast}$ & \textbf{65.5} & \textbf{40.0} & \textbf{59.0} & \textbf{57.1} & \textbf{55.4} \\ \bottomrule
\end{tabular}
\caption{MAB competency scores (\%), all using \model{GPT-4o-mini} for answer generation. Unstarred rows are published full-benchmark baselines from \cite{mab}. $^{\ast}$ReFind and Jetriever are evaluated head-to-head on the same dimension-proportional 60-question random subset (within AR, proportional sampling drew EventQA, MH-QA, and LME items, with no SH-QA item; dimension means aggregate the drawn subtasks).}\label{tab:mab}
\end{table}

\Needspace{12\baselineskip}\subsection{Disentangling world knowledge and score calibration}
\begin{figure}[ht]
\centering
\includegraphics[width=.95\linewidth]{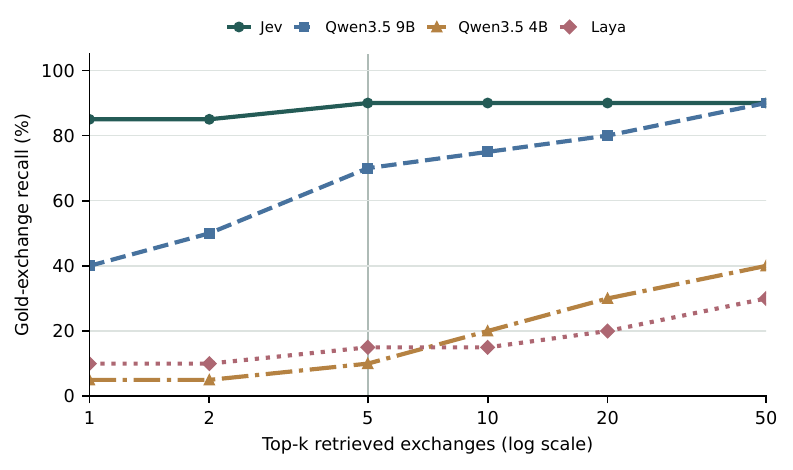}
\caption{Gold-exchange $\rat{k}$ at $k\in\{1,2,5,10,20,50\}$ across the same 20 InMind tasks (241 candidate exchanges per task; logarithmic horizontal axis). Markers reflect exact per-task gold ranks; \model{Qwen3.5-4B} \cite{qwen4b} is included under the same scoring protocol as \model{Qwen3.5-9B}.}\label{fig:inmind20-curve}
\end{figure}

To test Section~3's hypotheses directly, we compare representatives of all three scorer families on a shared 20-question InMind subset, with every model ranking the exact same 241 candidate exchanges per query:
\begin{itemize}\setlength{\itemsep}{2pt}\setlength{\parskip}{0pt}
  \item \textbf{BERT-based decision model (\model{Laya}):} a 421M ModernBERT-large encoder ({\small\nolinkurl{convaiinnovations/laya-typed-decisions}}) trained with a structured decision head \cite{laya,modernbert}.
  \item \textbf{Generative LLM (\model{Qwen3.5-9B}):} the post-trained {\small\texttt{Qwen/Qwen3.5-9B}} checkpoint \cite{qwen9b} (with \model{Qwen3.5-4B} \cite{qwen4b} also plotted in Figure~\ref{fig:inmind20-curve}), prompted for binary usefulness and scored by the normalized probability assigned to \texttt{true} versus \texttt{false}. Complete exchanges are scored directly whenever they fit within 1,200 tokens; only a single long background exchange per query (0.4\% of candidates, and none of the gold exchanges) exceeds this limit and is scored via 1,200-token windows (200-token overlap) with max-pooling.
  \item \textbf{System One decision model (\model{Jev}):} Jetriever's \texttt{Noul} utility scorer over each exchange.
\end{itemize}

\textbf{World knowledge governs broad recall.} Because InMind's gold exchanges share no lexical cues with the query, spotting their relevance at all requires pretrained world knowledge. At $\rat{50}$ in Figure~\ref{fig:inmind20-curve}, both LLM-scale models---Jev and Qwen3.5-9B---recover the gold exchange in 18 of 20 tasks (90\%), whereas the ModernBERT-based Laya recovers only 6 of 20 (30\%). Large-scale LLM pretraining supplies the factual breadth needed to recognize implicit connections; without that foundation, a BERT-scale decision model misses the majority of gold memories even across a 50-exchange shortlist.

\Needspace{7\baselineskip}\sidemark{READ THE ANALYSIS}{Please read the analysis on the left. This is the key distinction.}\textbf{Calibrated probabilities govern top-1 ranking.} Recognizing that the gold exchange matters somewhere in the top 50, however, is very different from ranking it first. Among 241 historical turns, plenty of background exchanges look loosely relevant or plausibly helpful to a user's question. Because Qwen3.5-9B's token probabilities are not calibrated to separate \emph{merely plausible} context from \emph{decisive} evidence, its recall falls sharply as $k$ tightens: 14/20 (70\%) at $\rat{5}$ and just 8/20 (40\%) at $\rat{1}$. Jev's RLCD-calibrated decision score cleanly isolates the decisive memory, achieving 17/20 (85\%) already at $\rat{1}$ and 18/20 (90\%) from $\rat{5}$ through $\rat{50}$. Both models possess the world knowledge to see why the gold exchange is relevant; probability calibration determines whether that exchange actually reaches the answerer at small $k$.

\Needspace{10\baselineskip}\subsection{Cost and throughput}\sidemark{AN ACCEPTABLE BILL}{Under a cent per InMind query, about three cents on MAB, and 6,446 pairs scored in 90 seconds. For full-pool retrieval, that is a very reasonable bill.}
Table~\ref{tab:cost} summarizes observed Jev API retrieval charges. Scoring all 241 exchanges per InMind query costs \$0.00834 on average (\$1.0425 across all 125 tasks). On the 60-question MAB sample, scoring an average of 1,258.2 conversations costs \$0.03103 per query (\$1.8617 total)---a mean heavily pulled up by one ReDial \cite{redial} task containing 42,946 candidate conversations, whereas the median MAB query costs just \$0.00730. Downstream \model{GPT-4o-mini} answer generation and judging across all 60 MAB tasks consumed 732,786 input and 1,472 output tokens, adding roughly \$0.1108 at standard API pricing \cite{gpt4omini-model} for a combined end-to-end run cost of \$1.97.

\begin{table}[ht]
\centering\small
\begin{tabular}{@{}lrrr@{}}\toprule
Run & Queries & Mean candidates & USD/query \\ \midrule
InMind exchanges & 125 & 241 & 0.00834 \\
InMind sessions & 125 & 47 & 0.00574 \\
MAB random conversations & 60 & 1,258.2 & 0.03103 \\ \bottomrule
\end{tabular}
\caption{Observed Jev API retrieval charges (excluding downstream answer generation). Total retrieval spend is \$1.0425 for InMind exchanges, \$0.7179 for InMind sessions, and \$1.8617 across 75,532 scored segments on MAB-60.}\label{tab:cost}
\end{table}

With up to 50 concurrent requests on InMind, individual candidate calls averaged 518~ms. In a separate batch throughput test over 10 LongMemEval queries \cite{longmemeval} totaling 6,446 query--exchange pairs, Jev scored the entire pool in 90.16~s at 100-way concurrency (71.5~scores/s) for \$0.247 in API charges. Because candidate judgments are mutually independent, request-level parallelism makes full-pool scoring genuinely practical at these memory sizes, though interactive single-query tail latency remains distinct from batch throughput.

\Needspace{14\baselineskip}\section{Discussion}
\subsection{Why System One decision models fit retrieval}
Section~3 argued that full-pool utility retrieval demands broad world knowledge, calibrated ranking probabilities, and low per-candidate cost---three properties tested empirically in Section~5 and summarized in Table~\ref{tab:model-families}. BERT-scale classifiers are cheap and produce sharp decision boundaries on narrow tasks, yet lack the open-world knowledge required when relevance hinges on unstated facts. Generative chat LLMs possess that world knowledge, but incur autoregressive overhead and produce uncalibrated token probabilities that struggle to push decisive evidence above plausible distractors. By distilling or adapting an LLM foundation into a non-generative, calibration-trained System One model, Jev hits the operating point where full-pool utility ranking works.

\begin{table}[ht]
\centering\footnotesize\setlength{\tabcolsep}{4pt}
\begin{tabular}{@{}lccc@{}}\toprule
 & \shortstack{Jev\\(System One model)} & \shortstack{Generative\\LLM scorer} & \shortstack{BERT-based\\classifier} \\ \midrule
Broad world knowledge & $\checkmark$ & $\checkmark$ & $\times$ \\
Calibrated ranking scores & $\checkmark$ & $\times$ & $\checkmark$ \\
Full-pool cost and latency & Low--moderate & High & Low \\ \bottomrule
\end{tabular}
\caption{Comparison of scorer families across the three requirements for full-pool utility retrieval.}\label{tab:model-families}
\end{table}

Stepping back, the three retrieval paradigms ask fundamentally different questions. Similarity search asks \emph{which stored text sounds like the query}. Agentic query expansion asks \emph{which hidden connection an LLM can guess before looking at the store}. Jetriever asks \emph{whether each stored memory actually changes how the query should be answered}. Whenever personal context connects to a user's request through everyday world knowledge, that third question is the right one to ask.

\subsection{Scaling limits}
Jetriever's obvious constraint is linear scaling: scoring $N$ memories takes $O(N)$ decisions per query. At the scales studied here---hundreds of exchanges in InMind and roughly a thousand conversations on average in MAB---high concurrency keeps retrieval within a few cents and seconds. For lifelong agents accumulating tens or hundreds of thousands of turns, however, naive per-pair API calls will eventually grow too costly and too slow.

\subsection{Toward lighter decision models with world knowledge}
When faced with $O(N)$ scaling, the reflexive retrieval-engineering instinct is to reintroduce a cheap BM25 or embedding pre-filter ahead of the scorer. For utility retrieval, however, that shortcut defeats the entire purpose of the system: any implicit memory whose connection lives in world knowledge will be silently discarded by a similarity first stage before the decision model ever sees it. Keeping full-pool utility scoring intact therefore points toward a more fundamental model-design challenge: \emph{building substantially lighter, faster System One decision models that still retain broad world knowledge}.

Figure~\ref{fig:inmind20-curve} illustrates the tension clearly: shrinking from \model{Qwen3.5-9B} to \model{Qwen3.5-4B} already erodes implicit recall, and dropping to the 421M-parameter \model{Laya} encoder loses most of it. Closing this efficiency--knowledge gap---through compact LLM-derived decision architectures, open-weight RLCD-style calibration training, and precomputed memory representations that amortize encoding across incoming queries---is the most promising path to scaling full-pool utility retrieval to lifelong agent memory.

\Needspace{10\baselineskip}\section{Conclusion}\sidemark{ONE LINE}{Find the memory that matters to the answer, not just one that sounds similar.}
Jetriever shifts agent memory retrieval from matching surface text to scoring downstream utility over the entire memory pool. Powered by Jev, this simple design recovers implicit-association memories on InMind that lexical, dense, and agentic memory baselines miss, improves multi-competency performance over ReFind on MemoryAgentBench, and keeps per-query retrieval costs in the cent range. Controlled comparisons further show why world knowledge alone is not enough: while both an ordinary LLM and Jev recognize implicit evidence within a wide shortlist, calibrated decision probabilities are what place the decisive memory at rank~1. Pushing full-pool utility scoring to much larger lifelong stores should not mean retreating behind similarity filters, but developing lighter, faster System One decision models that carry broad world knowledge at a fraction of today's compute.

\section*{Acknowledgments}
We thank GPT-6-Sol for assisting with the experimental runs---even though its draft prose was an unmitigated disaster---and Gemini 4 Carbon for helping rewrite and polish the manuscript.

\renewcommand{\refname}{\color{forest}References}
{\scriptsize\raggedright\setlength{\parskip}{0pt}
}

\clearpage
\appendix
\section{Full benchmark tables}
\subsection{InMind Memory Bench}\label{app:inmind-full}
\begin{table}[H]
\centering\small
\begin{tabular}{@{}lr@{}}\toprule
Configuration & Application (\%) \\ \midrule
\multicolumn{2}{@{}l}{\emph{Backbone control}}\\
GPT-5-mini, target in context & 84.0 \\
\midrule
\multicolumn{2}{@{}l}{\emph{Naive RAG}}\\
MiniLM & 9.6 \\
text-embedding-3-large & 16.0 \\
BM25 & 9.6 \\
\midrule
\multicolumn{2}{@{}l}{\emph{Memory systems with MiniLM}}\\
A-RAG & 4.8 \\
xMemory & 4.8 \\
Mem0 & 3.2 \\
A-Mem & 6.4 \\
HippoRAG 2 & 8.8 \\
MemoryOS & 8.0 \\
\midrule
\multicolumn{2}{@{}l}{\emph{Memory systems with text-embedding-3-large}}\\
A-RAG & 7.2 \\
xMemory & 6.4 \\
Mem0 & 6.4 \\
A-Mem & 9.6 \\
HippoRAG 2 & 5.6 \\
MemoryOS & 14.4 \\
\midrule
\multicolumn{2}{@{}l}{\emph{Jetriever with top-five exchanges}}\\
\rowcolor{pale}
Jetriever & \textbf{76.0} \\ \bottomrule
\end{tabular}
\caption{All 16 published InMind Application configurations \cite{inmind}, followed by Jetriever's top-five result. All rows use the same context-aware Application scoring criterion.}\label{tab:inmind-full}
\end{table}

\subsection{MemoryAgentBench}\label{app:mab-full}
\begin{table}[H]
\centering\scriptsize\setlength{\tabcolsep}{3pt}\renewcommand{\arraystretch}{.92}
\begin{tabular}{@{}lrrrrr@{}}\toprule
Original MAB baseline & AR & TTL & LRU & SF & Overall \\ \midrule
\multicolumn{6}{@{}l}{\emph{Long-context agents}}\\
GPT-4o (128K) & 58.1 & 50.0 & 54.9 & 32.5 & 48.8 \\
GPT-4o-mini (128K) & 49.2 & 48.6 & 46.2 & 25.0 & 42.2 \\
Claude-3.7-Sonnet (200K) & 59.7 & 53.9 & 62.2 & 22.5 & 49.6 \\
GPT-5-mini (400K) & 74.4 & 48.6 & 66.2 & 53.0 & 60.6 \\
GPT-4.1-mini (1M) & 71.8 & 46.2 & 49.1 & 20.5 & 46.9 \\
Gemini-2.0-Flash (1M) & 65.1 & 46.4 & 41.6 & 16.5 & 42.4 \\
GPT-4o-mini (reference) & 49.2 & 48.6 & 46.2 & 25.0 & 42.3 \\
\midrule
\multicolumn{6}{@{}l}{\emph{Simple RAG}}\\
BM25 & 60.5 & 44.5 & 35.6 & 25.5 & 41.5 \\
\midrule
\multicolumn{6}{@{}l}{\emph{Embedding RAG}}\\
Contriever & 33.9 & 42.9 & 29.8 & 12.5 & 29.8 \\
text-embedding-3-small & 53.8 & 42.7 & 36.3 & 15.5 & 37.1 \\
text-embedding-3-large & 54.6 & 44.3 & 37.3 & 16.0 & 38.0 \\
Qwen3-Embedding-4B & 54.7 & 45.1 & 37.0 & 16.0 & 38.2 \\
\midrule
\multicolumn{6}{@{}l}{\emph{Structure-augmented RAG}}\\
RAPTOR & 36.8 & 35.9 & 27.9 & 7.5 & 27.0 \\
GraphRAG & 40.9 & 24.8 & 19.9 & 8.0 & 23.4 \\
MemoRAG & 34.5 & 45.1 & 30.0 & 14.0 & 30.9 \\
HippoRAG 2 & 65.1 & 35.8 & 36.2 & 29.5 & 41.6 \\
Mem0 & 32.6 & 21.2 & 20.7 & 10.0 & 21.1 \\
Cognee & 28.3 & 22.8 & 16.0 & 15.5 & 20.6 \\
Zep & 37.5 & 37.5 & 16.2 & 5.0 & 24.0 \\
\midrule
\multicolumn{6}{@{}l}{\emph{Agentic memory}}\\
Self-RAG & 33.6 & 12.2 & 18.1 & 11.0 & 18.7 \\
MemGPT & 34.3 & 40.8 & 22.4 & 15.5 & 28.3 \\
MIRIX & 47.5 & 24.1 & 25.4 & 8.0 & 26.2 \\
MIRIX (4.1-mini) & 63.0 & 35.7 & 40.5 & 11.5 & 37.7 \\ \bottomrule
\end{tabular}
\caption{All 23 baselines in the original MAB main table \cite{mab}: four competency-group averages and Overall (\%). AR = Accurate Retrieval; TTL = Test-Time Learning; LRU = Long-Range Understanding; SF = Selective Forgetting. These are full-benchmark scores rather than MAB-60 subset scores.}\label{tab:mab-original}
\end{table}

\begin{table}[H]
\centering\scriptsize\setlength{\tabcolsep}{2.5pt}\renewcommand{\arraystretch}{.94}
\begin{tabular}{@{}lrrrrrrr@{}}\toprule
Published method & SH-QA & MH-QA & LME & EventQA & FC-SH & FC-MH & Avg. \\ \midrule
GPT-4o-mini & 64.0 & 43.0 & 30.7 & 59.0 & 45.0 & 5.0 & 41.1 \\
\midrule
BM25-RAG & 66.0 & 56.0 & 45.3 & 74.6 & 48.0 & 3.0 & 48.8 \\
Contriever & 22.0 & 31.0 & 15.7 & 66.8 & 18.0 & 7.0 & 26.8 \\
text-embed-3-small & 60.0 & 44.0 & 48.3 & 63.0 & 28.0 & 3.0 & 41.1 \\
text-embed-3-large & 54.0 & 44.0 & 50.3 & 70.0 & 28.0 & 4.0 & 41.7 \\
Qwen3-Embed-4B & 57.0 & 47.0 & 43.3 & 71.4 & 29.0 & 3.0 & 41.8 \\
\midrule
RAPTOR & 29.0 & 38.0 & 34.3 & 45.8 & 14.0 & 1.0 & 27.0 \\
GraphRAG & 47.0 & 47.0 & 35.0 & 34.4 & 14.0 & 2.0 & 29.9 \\
MemoRAG & 29.0 & 33.0 & 20.0 & 56.0 & 21.0 & 7.0 & 27.7 \\
HippoRAG 2 & 76.0 & 66.0 & 50.7 & 67.6 & 54.0 & 5.0 & 53.2 \\
Mem0 & 25.0 & 32.0 & 36.0 & 37.5 & 18.0 & 2.0 & 25.1 \\
Zep & 44.0 & 25.0 & 38.3 & 42.5 & 7.0 & 3.0 & 26.6 \\
\midrule
Self-RAG & 35.0 & 42.0 & 25.7 & 31.8 & 19.0 & 3.0 & 26.1 \\
MemGPT & 41.0 & 38.0 & 32.0 & 26.2 & 28.0 & 3.0 & 28.0 \\
MIRIX & 62.0 & 61.0 & 37.3 & 29.8 & 14.0 & 2.0 & 34.4 \\
ReFind & 83.0 & 69.0 & 51.3 & 74.1 & 62.7 & 8.8 & 58.2 \\ \bottomrule
\end{tabular}
\caption{Published full six-subtask comparison for Accurate Retrieval (SH-QA, MH-QA, LME, EventQA) and Selective Forgetting (FC-SH, FC-MH), reproduced from ReFind \cite{refind}, with reused baselines from MAB \cite{mab}. Avg.\ is the unweighted six-subtask mean (\%). This table reports full task sets rather than our 60-question subset.}\label{tab:mab-six-full}
\end{table}

\end{document}